\documentclass{article}

\usepackage[final]{timeseries_workshop}

\usepackage{natbib}
\usepackage{booktabs}  
\usepackage{graphicx}
\usepackage{caption}
\usepackage{subcaption}  
\usepackage[utf8]{inputenc} 
\usepackage[T1]{fontenc}    
\usepackage{hyperref}       
\usepackage{url}            
\usepackage{booktabs}       
\usepackage{amsfonts}       
\usepackage{nicefrac} 
\usepackage{pifont} 
\usepackage{microtype}      
\usepackage{xcolor}         
\usepackage{subcaption}
\usepackage{todonotes}
\usepackage{amssymb}
\usepackage{multirow}
\usepackage{booktabs} 
\usepackage{array}
\usepackage{authblk}

\title{How Foundational are Foundation Models for Time Series Forecasting?}

\author[1]{Nouha Karaouli}
\author[2]{Denis Coquenet}
\author[1]{Elisa Fromont}
\author[3]{Martial Mermillod}
\author[4]{Marina Reyboz}

\affil[1]{Univ. Rennes, CNRS, Inria, IRISA - UMR 6074, F-35000 Rennes, France}
\affil[2]{Univ. Rennes, CNRS, IRISA - UMR 6074, F-35000 Rennes, France}
\affil[3]{Univ. Grenoble Alpes, Univ. Savoie Mont Blanc, CNRS, LPNC, Grenoble, France}
\affil[4]{Univ. Grenoble Alpes, CEA, LIST, 38000 Grenoble, France}

\newcolumntype{C}[1]{>{\centering}m{#1}}

\begin{document}

\maketitle

\begin{abstract}
Foundation Models are designed to serve as versatile embedding machines, with strong zero shot capabilities and superior generalization performance when fine-tuned on diverse downstream tasks. While this is largely true for language and vision foundation models, we argue that the inherent diversity of time series data makes them less suited for building effective foundation models. We demonstrate this using forecasting as our downstream task. We show that the zero-shot capabilities of a time series foundation model are significantly influenced and tied to the specific domains it has been pretrained on. Furthermore, when applied to unseen real-world time series data, fine-tuned foundation models do not consistently yield substantially better results, relative to their increased parameter count and memory footprint, than smaller, dedicated models tailored to the specific forecasting task at hand.

\end{abstract}

\section{Introduction}

The emergence of Foundation Models (FMs), large-scale pretrained architectures such as BERT  \cite{devlin2019bert} in Natural Language Processing (NLP) and Vision Transformers \cite{dosovitskiy2020image} in Computer Vision (CV), has fundamentally transformed artificial intelligence. By leveraging massive and diverse datasets during pretraining, these models exhibit strong generalization abilities, enabling zero-shot and few-shot transfer to a wide range of downstream tasks \cite{brown2020language, radford2021learning}. This shift has allowed FMs to consistently outperform traditional task-specific models trained from scratch for narrowly defined problems \cite{radford2021learning}.

Inspired by these successes, researchers have recently proposed Time Series Foundation Models (TSFMs), large pretrained models designed to capture general-purpose representations across diverse temporal data \cite{liang2024foundation}. These models aim to transfer knowledge across forecasting tasks by learning temporal patterns at scale, showing promising results in a variety of domains with minimal task-specific tuning.

However, the time series domain poses unique challenges that set it apart from NLP and CV. Time series data often exhibits domain-specific structures such as seasonality, trends, irregular sampling, and high variability across applications, even within the same broad category \cite{ye2024survey}. Such characteristics introduce distribution shifts that undermine the generalization abilities of TSFMs \cite{li2024foundts}. In particular, our experiments suggest that TSFMs’ zero-shot performance is highly sensitive to the alignment between the statistical properties of the pretraining and target domains. When this alignment is weak, we observe substantial drops in generalization, even across domains that might appear related.

While TSFMs often benefit from rapid initial convergence, extended fine‑tuning can lead to performance degradation, whereas task‑specific models trained from scratch typically yield steady accuracy gains under longer training and limited data regimes \cite{zhao2025moreunlockingspecializationtime}.

Motivated by these challenges, we conduct a thorough empirical evaluation of the univariate forecasting capabilities of TSFMs across diverse tasks. We compare them with traditional models trained from scratch to assess whether TSFMs offer practical advantages when fine-tuned on specific, potentially domain-shifted datasets.

Our main contributions are:
\begin{itemize}
    \item Evaluating TSFMs in zero-shot mode across both domain-related and domain-shifted forecasting datasets.
    \item Comparing the fine-tuning capabilities of TSFMs versus traditional models on forecasting tasks to evaluate their adaptability and effectiveness under domain shift and limited data.
    \item Proposing a new forecasting dataset consisting of daily electricity usage over two years, on which a small dedicated network achieves better results than a fine-tuned TSFM.
\end{itemize}

\section{Related Work}
Recent TSFMs such as TiReX \cite{auer2025tirex}, TimeGPT \citep{garza2024timegpt}, TimesFM \citep{das2024timesfm}, and FEDformer \citep{pmlr-v162-zhou22g} leverage large-scale pretraining to enable strong generalization and transfer across forecasting tasks.

To assess their practical utility, several benchmarking frameworks have emerged. GIFT-eval \cite{aksu2024gift} measures cross-domain generalization using standardized protocols, OpenTS \cite{opents2024} offers a reproducible suite spanning datasets, horizons, and metrics, while Nixtla’s Arena \cite{nixtla2024arena} provides a comprehensive evaluation across frequencies and domains. \cite{xu2025specialized} have also pointed out that naive baselines (here, a simple auto-regressive model) can achieve competitive performance compared to TSFM on several forecasting tasks.

These efforts report promising performance on public datasets such as Monash \cite{godahewa2021monash} and ETT \cite{zhou2021informer}. However, we had to compare the generalization performance of these foundation models on time series ensured to be completely new and not included in these benchmark databases in order to  test the challenges faced in deployment.

In contrast, we evaluate TSFMs on a proprietary electricity consumption dataset with realistic and complex domain shifts not seen during pretraining. Our setup introduces explicit distributional changes, enabling a more rigorous assessment of generalization. 

Contrary to standard benchmarks that primarily focus on evaluating zero-shot capabilities of TSFMs on public datasets, we further compare these models to conventional ones trained from scratch. This allows us to highlight scenarios where smaller, specialized models achieve comparable performance to large pretrained TSFMs, especially under conditions of data scarcity and nonstationarity.

Through this, we uncover limitations in TSFMs' robustness and provide new insights into their practical effectiveness in real-world forecasting scenarios.

\section{Methodology}

Our evaluation addresses two central questions: (1) Can TSFMs generalize beyond their pretraining distributions? (2) Are they practically competitive with lightweight, specialized alternatives?

We benchmark three leading TSFMs, namely TimesFM~\citep{das2024timesfm}, TimeGPT~\citep{garza2024timegpt}, and TiReX~\citep{auer2025tirex}, alongside SAMFormer \cite{ilbert2024samformer}, a compact attention-based model operating over the channel dimension. Unlike the other models, SAMFormer is trained from scratch in our experiments.

\vspace{0.5em}
\textbf{Synthetic benchmarks.} We construct four datasets that reflect recurring structures in TSFM pretraining, while ensuring zero data overlap. 

\begin{itemize}
    \item \textbf{D1} and \textbf{D2} are composed of harmonically aligned sine waves with full observability, probing the models' ability to recognize and extrapolate clean periodic signals.
    
    \item \textbf{D3} and \textbf{D4} consist of randomly sampled, non-harmonic sine waves, forming complex, partially observable cycles. These challenge the models to generalize from incomplete patterns.
\end{itemize}

All synthetic sequences contain 2,688 time steps (8 weeks sampled at 30-minute intervals).

\vspace{0.5em}
\textbf{Real-world evaluation.} We test TSFMs on \textit{Elec\_Consumption}, a proprietary small dataset capturing daily electricity usage of a single home over two years (2023–2024). Unlike the generic, population-level datasets typically used during TSFM pretraining, this dataset reflects individual consumption behavior, introducing a clear distribution shift. This setting allows us to rigorously evaluate whether pretrained models retain strong forecasting performance when faced with personalized, unseen patterns, a crucial requirement for real-world deployment in user-specific applications.

All datasets used in this work are made publicly available for reproducibility\footnote{\url{https://doi.org/10.5281/zenodo.17199246}}.

\vspace{0.5em}
\textbf{Fine-tuning experiments.} We fine-tune TimesFM on Elec\_Consumption and compare it to SAMFormer trained from scratch. This setup quantifies the trade-off between the computational overhead of fine-tuning large pretrained models and the efficiency of smaller models tailored to specific domains.

Together, these evaluations dissect the \textit{one-size-fits-all} \citep{bommasani2021opportunities, yuan2025beyond}
promise of TSFMs, distinguishing their theoretical representational capacity (via synthetic benchmarks) from their practical effectiveness in real-world deployment. We report Mean Absolute Error (MAE) as the primary metric.

\section{Results}

We begin our experimental evaluation by testing all models in zero-shot mode on both synthetic and real-world datasets, Appendix~\ref{appendix:Querying FM} provides additional details on our models querying setup. Tables~\ref{tab:synthetic_d1_d2} and~\ref{tab:synthetic_d3_d4} report results on synthetic data using a fixed context length of 512 across three forecast horizons. Table~\ref{tab:real_data_d1_context_horizon} presents results on the Elec\_Consumption dataset.

\begin{table}[h]
\centering
\caption{Zero-shot MAE on  D1 and D2 for various forecasting horizons and models. Lower is better.}
\label{tab:synthetic_d1_d2}
\begin{tabular}{lp{1cm}  ccc  ccc}
\toprule
\multicolumn{2}{c}{Datasets} & \multicolumn{3}{c}{D1} & \multicolumn{3}{c}{D2} \\
\cmidrule(lr){3-5}
\cmidrule(lr){6-8}
\multicolumn{2}{c}{Models} & TimeGPT & TiReX & TimesFM & TimeGPT & TiReX & TimesFM \\
\midrule
\multirow{3}{*}{Horizons}  & \textbf{128} & 0.89 & \textbf{0.11}& 0.13 & 0.80 & 0.29 & \textbf{0.15} \\
        & \textbf{256} & 1.08& \textbf{0.21}& 0.22 & 1.25 & 0.72 & \textbf{0.35} \\        & \textbf{512} & 1.09 & 0.37& \textbf{0.34} & 1.57 & 1.11 & \textbf{0.72} \\
\bottomrule
\end{tabular}
\end{table}

\begin{table}[h]
\centering
\caption{Zero-shot MAE on  D3 and D4 for various forecasting horizons and models. Lower is better.}
\label{tab:synthetic_d3_d4}
\begin{tabular}{lp{1cm} ccc  ccc}
\toprule
\multicolumn{2}{c}{Datasets} & \multicolumn{3}{c}{D3} & \multicolumn{3}{c}{D4} \\
\cmidrule(lr){3-5}
\cmidrule(lr){6-8}
\multicolumn{2}{c}{Models} & TimeGPT & TiReX & TimesFM & TimeGPT & TiReX & TimesFM \\
\midrule
\multirow{3}{*}{Horizons}  & \textbf{128} & 1.86 & \textbf{1.1} & 1.13 & 1.3 & \textbf{0.78} & 0.89  \\
        & \textbf{256}  &  1.43 & \textbf{0.95} & 0.98 & 1.63 & \textbf{1.6} & 1.62 \\
        & \textbf{512}  & \textbf{2.29} & 3.3  & 3.5 & \textbf{2.31} & 2.8 & 2.98 \\
\bottomrule
\end{tabular}
\end{table}

\begin{table}[h]
\centering
\caption{Zero-shot MAE on \textit{Elec\_Consumption} for varying context-horizon pairs and models. Lower is better.}
\label{tab:real_data_d1_context_horizon}
\setlength{\tabcolsep}{10pt} 
\begin{tabular}{lccccc}
\toprule
\multirow{2}{*}{Models}
& \multicolumn{5}{c}{Context -- Horizon}\\
\cmidrule(lr){2-6}
& 15--7 & 30--7 & 60--30 & 128--128 & 365--365 \\
\midrule
TimeGPT  & 6.60 & 6.52 & 5.60 & 6.91 & 6.44 \\
TiReX    & 6.94 & \textbf{5.71} & 4.61 & \textbf{3.78} & 5.90 \\
TimesFM  & \textbf{5.07} & 5.83 & \textbf{4.08} & 4.63 & \textbf{5.30} \\
\bottomrule
\end{tabular}
\end{table}

Among all five experiments, TiReX and TimesFM consistently perform best, particularly on D1 and D2, which exhibit simple and periodic sinusoidal patterns, highlighting their ability to capture repetitive temporal structures. In contrast, forecasting on D3 and D4, involving irregular and composite sinusoidal signals, is more challenging. Despite this, foundation models still generalize reasonably well, likely due to pretraining on structurally similar synthetic patterns. However, on the real-world Elec\_consumption dataset, even with careful tuning of context and horizon lengths, the models struggle to accurately forecast the future values. This shows the limits of the generalization abilities of current state-of-the-art TSFMs for a real-case forecasting scenario.

\begin{figure}[htbp]
    \centering
    \begin{subfigure}{0.95\linewidth}
        \centering
        \includegraphics[width=\linewidth]{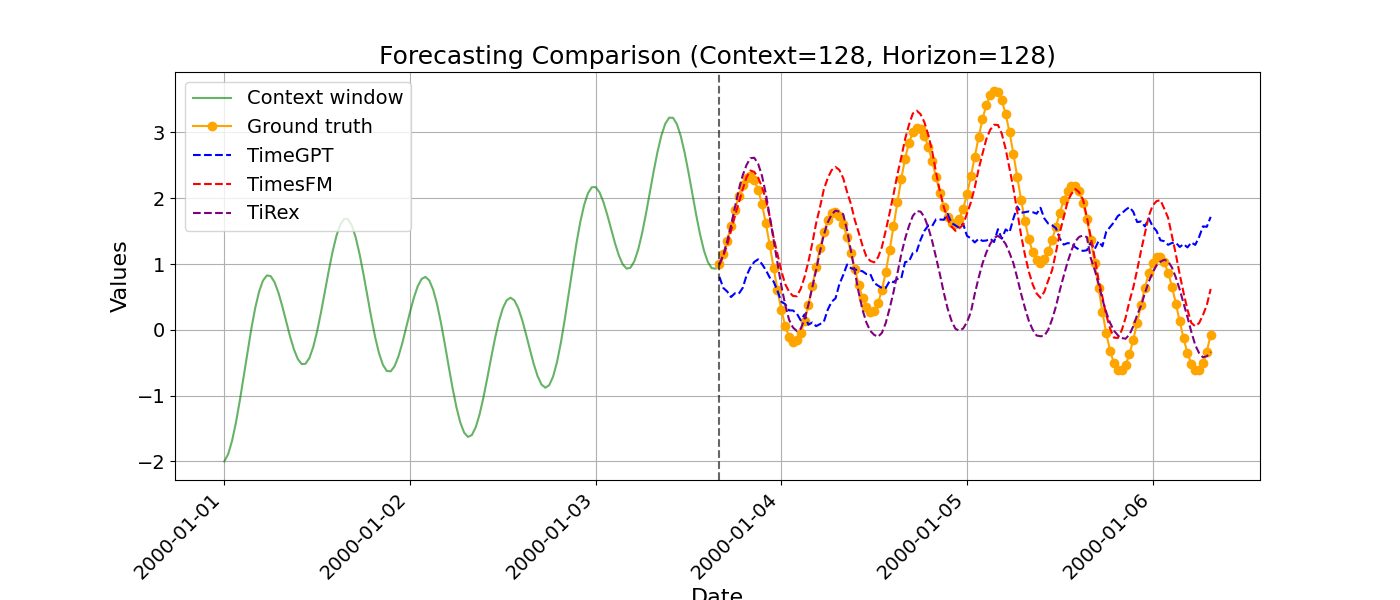}
        \caption{Forecasting results on D1.}
        \label{fig:results_d1}
    \end{subfigure}
    
    \vspace{0.5em} 

    \begin{subfigure}{0.95\linewidth}
        \centering
        \includegraphics[width=\linewidth]{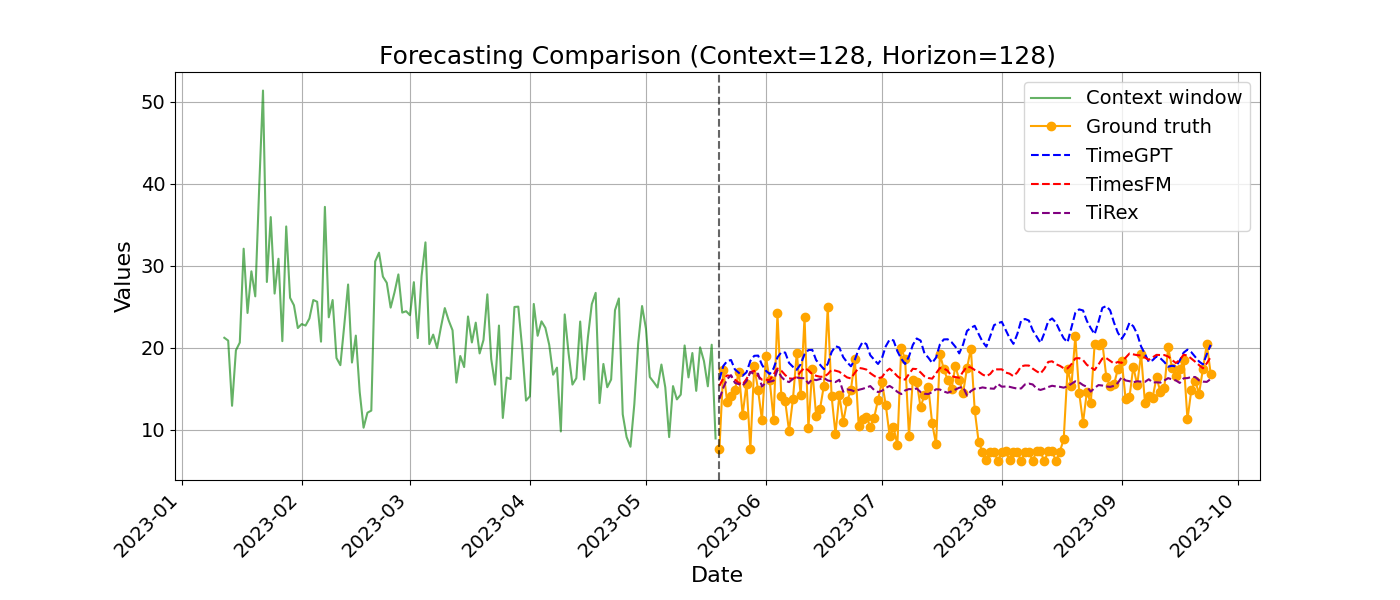}
        \caption{Forecasting results on Elec\_Consumption.}
        \label{fig:results_elec}
    \end{subfigure}

    \caption{Forecasting results using different models. Panel (a) shows results on D1, while panel (b) presents results on Elec\_Consumption.}
    \label{fig:results_d1_elec}
\end{figure}

This performance contrast is clearly illustrated in Figure \ref{fig:results_d1_elec}. On dataset D1, the TSFMs demonstrate strong generalization, achieving MAE scores of 0.95, 0.46, and 0.71 on TimesGPT, TimesFM, and TiReX, respectively. In contrast, on Elec\_consumption, the forecasts deviate more noticeably from the expected values.

While zero-shot results show that TSFMs perform well on target distributions that resemble their pretraining data, their ability to adapt to small, domain-specific datasets produces high errors and low prediction ability. To investigate this, we compare fine-tuned TimesFM with SAMFormer trained from scratch on our Elec\_consumption dataset. This evaluation tests whether TSFMs’ learned representations and inductive biases confer advantages for personalization.

Fine-tuning and training from scratch were performed using Adam with a learning rate $10^{-4}$, weight decay $0.01$, and batch size $64$. The choice of LR follows the default fine-tuning configuration used in the public TimesFM examples, ensuring consistency with recommended practice for this foundation model. Data were standardized and framed with a sliding window (context $=128$, horizon $=128$). TimesFM was fine-tuned from a fixed pre-trained checkpoint, excluding any significant source of randomness. In contrast, SAMFormer was trained from scratch, introducing natural variability in the results due to the random weight initialization. To make the evaluation robust, we computed the mean and standard deviation over $5$ runs with different random seeds. Models were trained for up to $100$ epochs with early stopping (patience $=10$). Experiments were conducted on an NVIDIA Tesla V100 GPU. Results are shown in Table \ref{tab:mae_lr_single}.

\renewcommand{\arraystretch}{1.2}
\begin{table}[h]
\centering
\caption{MAE for TimesFM and SAMFormer with a context window of 128 and a forecast
horizon of 128.}
\label{tab:mae_lr_single}
\begin{tabular}{lc}
\toprule
Models & MAE \\
\midrule
TimesFM   & $4.49 \pm 0.00$\\
SAMFormer & \textbf{4.28 ± 0.05} \\

\bottomrule
\end{tabular}
\end{table}

As one can note, the results show that SAMFormer, trained entirely from scratch with fewer than 50K parameters, ultimately outperforms TimesFM on the target forecasting task. While TimesFM benefits from large-scale pretraining and contains over 500 million parameters, SAMFormer achieves superior accuracy while remaining extremely lightweight and efficient to train on consumer-grade GPUs. This contrast highlights a key point: massive pretrained models do not always guarantee superior downstream performance, particularly in settings where data distributions differ from the pretraining corpus or where the target domain exhibits specific structural regularities that a smaller model can exploit more effectively. Moreover, SAMFormer’s compact size reduces both training time and inference cost, making it well-suited for rapid experimentation and deployment in resource-constrained environments. These findings illustrate that carefully designed, domain-adapted models can deliver competitive or even superior performance compared to large foundation models, while offering substantial advantages in efficiency, accessibility, and environmental sustainability.

\section{Conclusion}
While TSFMs show strong zero-shot performance on synthetic and structurally similar data, their generalization ability is tightly coupled with the distribution seen during pretraining. In real-world settings involving domain shifts and limited data, a lightweight model like SAMFormer, with only 49.5K parameters and no large-scale pretraining, can still achieve better results when trained from scratch. This suggests that the “one-size-fits-all” promise of TSFMs may not hold in practice, especially under resource constraints or personalization requirements. Our findings advocate for a more nuanced deployment strategy: leveraging TSFMs when pretraining-task similarity is high, and favoring lightweight, specialized models when personalization, efficiency, or domain mismatch is critical.

\bibliographystyle{unsrt}

\newpage
\appendix
\section{Querying Foundation Models}
\label{appendix:Querying FM}

We evaluated three pretrained time-series foundation models in a zero-shot setting: TimesFM \citep{das2024timesfm}, TiReX \citep{auer2025tirex}, and TimeGPT \citep{garza2024timegpt}. All models were used without architectural modifications; only the context length, forecast horizon, and sampling frequency were adapted to match our datasets. This ensures comparability while respecting each model's inference interface.

\paragraph{TimesFM.} 
TimesFM is a decoder-only transformer trained on large-scale synthetic and real-world collections of time series. The model represents inputs as patches of fixed length embedded into a 1280-dimensional latent space, with autoregressive decoding over output patches. We employed the released checkpoint \texttt{google/timesfm-2.0-500m-pytorch} and queried the model with context windows of varying length ($L$) and horizons matched to the experimental setup ($H$). Forecasts were generated via the \texttt{forecast\_on\_df} API, with the frequency parameter aligned to the dataset resolution (30 minutes for synthetic benchmarks; daily for \textit{Elec\_Consumption}).

\paragraph{TiReX.} 
TiRex is a TSFM built upon the xLSTM architecture. The pretrained checkpoint \texttt{NX-AI/TiReX} was queried directly, providing context windows of length $L$ and requesting horizons of length $H$. The model outputs both quantile forecasts and mean trajectories; we report the mean predictions in all experiments. Dataset frequency was set to match the native resolution, consistent with the TimesFM setup.

\paragraph{TimeGPT.} 
TimeGPT is a hosted transformer-based FM trained on over 100B time-series observations across diverse domains. We accessed the \texttt{timegpt-1-long-horizon} variant through the Nixtla Python client, using an API key for authentication. The model was queried with training windows of length $L$ and a forecast horizon $H$ identical to the experimental split. Frequency alignment was handled through the timestamp column in the input dataframe.

Across all models, context length ($L$), horizon length ($H$), and sampling frequency were varied according to the dataset and experimental condition. No fine-tuning or weight adaptation was applied; results therefore reflect pure zero-shot performance under consistent querying protocols.

\end{document}